# Predicting Targeted Therapy Resistance in Non-Small Cell Lung Cancer Using Multimodal Machine Learning

Peiying Hua, MS[1]; Andrea Olofson, MD[2]; Faraz Farhadi, MD[3]; Liesbeth Hondelink, MD[4]; Gregory Tsongalis, PhD[5]; Konstantin Dragnev, MD[6]; Dagmar Hoegemann Savellano, MD[7]; Arief Suriawinata, MD[5]; Laura Tafe, MD[5]; Saeed Hassanpour, PhD[1,8,9*]

[1]Department of Biomedical Data Science, Geisel School of Medicine at Dartmouth, Hanover, NH 03755, USA

[2]Department of Pathology, Ochsner Health System, New Orleans, LA 70115, USA

[3]Geisel School of Medicine at Dartmouth, Hanover, NH 03755, USA

[4]Department of Pathology, Leiden University Medical Center, Leiden, the Netherlands

[5]Department of Pathology and Laboratory Medicine, Dartmouth-Hitchcock Medical Center, Lebanon, NH 03756, USA

[6]Department of Medical Oncology, Dartmouth-Hitchcock Medical Center, Lebanon, NH 03756, USA

[7]Department of Radiology, Dartmouth-Hitchcock Medical Center, Lebanon, NH 03756, USA

[8]Department of Epidemiology, Geisel School of Medicine at Dartmouth, Hanover, NH 03755, USA

[9]Department of Computer Science, Dartmouth College, Hanover, NH 03755, USA

**Corresponding author:** Saeed Hassanpour, PhD (Email: Saeed.Hassanpour@Dartmouth.edu)

**Running title:** Predicting Resistance in Lung Cancer with Multimodal AI

**Word count:** 3189

**Number of figures, videos, and tables:** 9

**Contributions:**

(I) Conception and design: S Hassanpour

(II) Administrative support: S Hassanpour

(III) Provision of study materials or patients: A Olofson, F Farhadi, L Hondelink, A Suriawinata

(IV) Collection and assembly of data: A Olofson, F Farhadi, L Hondelink, A Suriawinata

(V) Data analysis and interpretation: P Hua

(VI) Manuscript writing: P Hua, S Hassanpour

(VII) Final approval of manuscript: All authors

## Abstract

**Background:** Resistance to tyrosine kinase inhibitors remains a major clinical challenge in the treatment of non-small cell lung cancer (NSCLC) with activating epidermal growth factor receptor (EGFR) mutations. Despite the efficacy of third-generation EGFR inhibitors, no standard tool currently exists to predict resistance using routinely available clinical data.

**Methods:** We conducted a multi-institutional retrospective study to develop and evaluate a multimodal machine learning model for predicting therapy resistance in late-stage NSCLC patients with EGFR mutations. The study included 42 patients treated with EGFR-targeted therapy from Dartmouth Hitchcock Medical Center and Ochsner Health, using data including histology whole-slide images, next-generation sequencing results, and demographic and clinical variables. The modeling framework fused image and non-image data through a three-stage training process and was evaluated using 5-fold nested cross-validation. Model performance was assessed using the concordance index (c-index), Kaplan-Meier survival curves, and log-rank tests. Interpretability analyses were conducted using attention maps, feature importance coefficients, and cellular composition comparisons.

**Results:** The multimodal model achieved a mean c-index of 0.82 across cross-validation folds, outperforming image-only and non-image models (c-index 0.75 and 0.77, respectively). Stratified analyses across institutions confirmed consistent performance gains with the multimodal approach. Kaplan-Meier analysis revealed that the multimodal model significantly stratified patients into distinct hazard groups (log-rank $P = 0.04$), which unimodal models failed to achieve. Key predictors included RB1 mutation and Hispanic ethnicity. Attention maps highlighted histologic regions with deformed nuclei, and cellular analysis revealed reduced inflammatory cell presence in high-risk patients.

**Conclusions:** This study presents a robust multimodal machine learning model for predicting therapy resistance in EGFR-mutant NSCLC, leveraging routinely collected clinical data without manual feature engineering. The model demonstrated superior performance over unimodal models and effective hazard stratification, suggesting utility for personalized treatment decisions. These findings underscore the potential of multimodal AI tools to advance precision oncology, particularly in resource-limited settings. Further validation in larger, diverse cohorts is warranted.

**Key findings**

- A multimodal machine learning model was developed to predict resistance in late-stage non-small cell lung cancer (NSCLC) patients with EGFR mutations.
- By integrating histology images, genomic alterations, and clinical data, the model achieved a mean c-index of 0.82 and outperformed unimodal models in predictive accuracy and hazard stratification.
- The model used routinely collected clinical data and required no manual feature engineering.

**What is known and what is new?**
- *Known*: Resistance to EGFR-targeted therapies remains a major challenge in NSCLC, and no validated tools exist to predict resistance using multimodal clinical data.
- *New*: This study introduces a multimodal machine learning framework that combines histopathology, genomics, and clinical data to predict therapeutic resistance and stratify patient risk, showing consistent improvements over unimodal approaches.

**What is the implication, and what should change now?**
- This model can support personalized treatment planning and improve prognostic accuracy using existing patient data.
- Future studies should validate this approach in larger, diverse cohorts to enable clinical adoption.

# 1. Introduction

Lung cancer remains the leading cause of cancer-related deaths worldwide, accounting for approximately 1.6 million deaths annually, or nearly one-quarter of all cancer mortalities (1). Non-small cell lung cancer (NSCLC) constitutes approximately 85% of all lung cancer cases (2), and among these, a significant proportion (approximately 32%) harbor activating mutations in the epidermal growth factor receptor (EGFR) gene (3). Targeted therapies such as osimertinib, a third-generation EGFR tyrosine kinase inhibitor (TKI), have demonstrated improved progression-free and overall survival in patients with EGFR-mutated NSCLC (4,5). Despite its efficacy, many patients inevitably develop resistance to osimertinib, often within 10–19 months of treatment initiation (6), and are left with limited therapeutic alternatives beyond chemotherapy or local ablative strategies (7,8).

Currently, no standardized clinical tools exist to predict treatment resistance in EGFR-mutant NSCLC. This limits the ability to personalize treatment strategies and preemptively adjust care plans. The lack of predictive tools also contributes to uncertainty in patient counseling and adds financial and emotional burdens, particularly for patients facing the high cost and potential toxicities of targeted therapies like osimertinib (9,10). Meanwhile, multimodal data—including histopathology slides, next-generation sequencing (NGS), and clinical variables—are routinely collected in oncology practice but remain underutilized in predicting therapeutic response.

Recent advances in machine learning and artificial intelligence offer new opportunities to harness such multimodal data for predictive modeling in oncology. Integrating disparate data sources through machine learning models has shown promise in improving prognostic accuracy and treatment stratification (11–14). However, few studies have evaluated multimodal models for resistance prediction specifically in the context of EGFR-targeted therapy.

To address this gap, we conducted a multi-institutional retrospective study to develop and evaluate a multimodal machine learning model that integrates histology, genomics, and clinical data to predict resistance in late-stage EGFR-mutant NSCLC patients. Our objective was to create a clinically interpretable, non-invasive tool using routinely collected patient data to support personalized treatment decisions and enhance precision oncology efforts. This manuscript is written following the STROBE checklist.

## 2. Methods

### 2.1 Datasets

This retrospective study analyzed data from 42 patients with stage IIIb or IV NSCLC harboring EGFR-activating mutations, treated with osimertinib at Dartmouth Hitchcock Medical Center (DHMC, n=23) and Ochsner Health (n=19). Inclusion criteria required patients to have received osimertinib as first- or later-line therapy and to have undergone next-generation sequencing (NGS) confirming EGFR mutations. Patients with carcinoma of unknown primary, non-NSCLC histology, use of osimertinib in adjuvant or neoadjuvant settings, or presence of de novo resistance mutations were excluded. Collected data included demographic variables, prior treatments, pathology reports, histological whole slide images, NGS profiles, radiology reports, and clinical outcomes. The utilization of data was approved by the institutional review board (IRB) at each institution.

### 2.2 Data pre-processing

NGS data were processed into binary representations of mutation status. Furthermore, certain mutations of potential clinical importance according to the literature were identified. These include the L858R and T790M mutations in the EGFR gene, EGFR exon 19 deletion, EGFR amplification, and disruptive TP53 mutation (15). Histopathology slides were scanned at 20× (0.5 μm/pixel) or 40× (0.25 μm/pixel) magnification using Leica AT2 and Philips UltraFast scanners.

### 2.3 Single Modality Models

#### 2.3.1 Non-Image Modality

For the non-image modality, we modeled osimertinib resistance using next-generation sequencing (NGS) data, demographic information, and other clinical variables. Given the limited number of eligible NSCLC patients treated with osimertinib, the available sample size was small. We selected Cox proportional hazards regression as the primary model due to its interpretability, established utility in survival analysis, and reliability in low-sample contexts (16,17).

To evaluate the effect of model architecture on overall multimodal performance, we conducted an experiment comparing Cox regression with two alternative approaches: self-normalizing neural networks (SNNs) and multilayer perceptrons (MLPs) (18). SNNs, introduced by Klambauer et al. (19), maintain stable activations through scaled exponential linear units (SELUs) and are well suited for high-dimensional, small-sample scenarios. Both SNN and MLP models were implemented using compact two-layer architectures with eight neurons per layer and trained with L1/L2 regularization and dropout to reduce overfitting. This comparative analysis allowed us to assess the trade-offs between model simplicity, interpretability, and flexibility in representing non-linear relationships within the non-image modality.

### 2.3.2 Image Modality

For the image modality, we implemented a two-stage deep learning pipeline comprising a convolutional feature extractor followed by a transformer-based aggregator to model spatial patterns predictive of therapy resistance. Whole slide histology images were first preprocessed and tiled into non-overlapping patches. Patch-level features were extracted using a ResNet-18 model pretrained on lung adenocarcinoma histopathology images (20). These features served as input to a vision transformer (ViT) architecture consisting of 12 transformer encoder layers with 8-head self-attention modules per layer (21).

The vision transformer was initialized with weights pretrained on The Cancer Genome Atlas (TCGA) (22) histology slides from five cancer types to provide contextual understanding across varied tissue morphologies (23). Fine-tuning was then performed using the study-specific histology data to adapt the model to the resistance prediction task. The self-attention mechanism in the transformer allowed the model to capture both local and global tissue-level dependencies, enhancing its ability to identify subtle but predictive histopathological cues. The final output from the transformer was a 32-dimensional feature embedding for each patient, which was passed to the fusion layer in the multimodal model. This image-based feature extraction strategy enabled high-capacity representation of histologic complexity while maintaining modularity for integration with other clinical and genomic data streams.

### 2.3.3 Fusion Layer

In our multimodal framework, each unimodal model—the image and non-image branches—was first trained independently to serve as a dedicated feature extractor. The resulting feature vectors were then combined in a fusion layer for outcome prediction using a Cox proportional hazards model (**Figure 1**). We systematically evaluated multiple fusion strategies to determine the most effective method for integrating information across modalities. In the early fusion strategy, intermediate latent features extracted prior to the final prediction layer in each unimodal model were concatenated and passed to the fusion model. In the late fusion strategy, the predicted risk scores from each unimodal output layer were combined as input to the final Cox regression model. Both approaches were evaluated under various configurations, yielding four distinct fusion scenarios.

Model performance under each fusion strategy was assessed using the c-index across 5-fold nested cross-validation. This design ensured unbiased estimation and helped identify the optimal integration approach for the multimodal pipeline. The best-performing configuration was retained for final evaluation. By treating each modality as a complementary information source and explicitly testing multiple fusion mechanisms, this strategy enabled us to identify a robust and interpretable method for combining heterogeneous patient data streams.

### 2.3.4 Loss Function

To accommodate the time-to-event nature of our primary outcome—progression-free survival—we employed a negative log partial likelihood function derived from the Cox proportional hazards model as the loss function for training. This formulation enables the model to learn risk scores associated with covariate patterns while appropriately handling censoring in survival data. The loss function is defined as follows:

$$Loss(\theta) = -\frac{1}{N_{event}} \sum_{i:event} (\hat{h}_\theta(x_i) - \log \sum_{j \in R(T_i)} e^{\hat{h}_\theta(x_j)})$$

Where $N_{event}$ is the total number of patients who developed resistance during the observation period, $\hat{h}_\theta(x_i)$ denotes the predicted risk score for patient $i$ given model parameters $\theta$, and $R(T_i)$ represents the risk set, i.e., the group of patients still at risk at time $T_i$, the event time for patient $i$. This approach allows for effective optimization of survival risk rankings without requiring specification of the baseline hazard function. It also facilitates direct comparison of multimodal model performance across validation folds via concordance-based metrics.

### 2.4 Multi-Model Training and Evaluation

The multimodal model was trained using a three-stage process to optimize feature extraction and mitigate overfitting given the limited sample size. First, the non-image modality was trained using demographic, NGS, and clinical data. Next, the image modality was trained using histology slides. Finally, the fusion layer was trained using extracted features from both modalities. This modular training strategy decomposed the multimodal pipeline into simpler components, enabling more stable and effective training with limited data.

To enhance generalization and reduce overfitting, we applied regularization and dropout techniques. A 5-fold nested cross-validation was used to prevent information leakage, ensuring consistent training, validation, and testing splits. Model performance was evaluated using the c-index, with 95% confidence intervals to quantify uncertainty.

**2.5 Model Interpretation and visualization**

To enhance interpretability of the multimodal model, we analyzed the contributions of both non-image features and spatial patterns in histology slides. For the non-image branch, we used the coefficients from the Cox regression model to quantify feature importance, interpreting the magnitude and direction of each variable's association with predicted resistance risk. For the image modality, we generated attention heatmaps by averaging self-attention weights across heads and applying recursive multiplication, enabling spatial localization of histologic regions contributing most strongly to model predictions (21). We compared attention distributions before and after fine-tuning to identify shifts in model focus during domain adaptation. To further characterize histologic correlates of risk, we applied a pretrained Hover-Net model (24) to segment and classify nuclei into tumor, inflammatory, stromal, and other cell types. This allowed us to compare cellular composition between predicted high- and low-risk groups, offering biological insight into differential model predictions.

**3. Results**

**3.1 Study samples**

The final cohort included 42 patients with stage IIIb or IV NSCLC harboring activating EGFR mutations. At the time of last follow-up, 14 patients (33.3%) had experienced disease recurrence. The mean age was 68.8 years (SD: 11.9), and the majority were female (71.4%). Of the cohort, 54.8% were never-smokers and 7.1% were current smokers. Each patient had an average of 2.0 (SD: 1.2) mutations among the selected hotspot genes, including the EGFR mutation required for inclusion. A detailed summary of patient characteristics stratified by clinical site is provided in **Table 1**.

**3.2 Model performance and comparison between single and multiple modality models**

Across 5-fold nested cross-validation, the multimodal model integrating image and non-image features achieved a mean c-index of 0.82 (SD: 0.17; 95% CI: 0.62–1.00), outperforming unimodal models based on non-image (mean c-index: 0.77) and image-only (mean c-index: 0.75) data (**Table 2**). Pairwise t-tests across folds indicated that the multimodal model improved c-index over non-image and image models by 0.06 (p = 0.25) and 0.07 (p = 0.17), respectively. Compared with random prediction (c-index = 0.5), only the multimodal model showed a statistically significant improvement (p = 0.01), whereas unimodal models did not reach significance (p = 0.06 for non-image; p = 0.08 for image).

Stratified site-level analysis showed consistent results. Among DHMC patients, the multimodal model achieved a c-index of 0.78, compared to 0.73 and 0.75 for non-image and image modalities, respectively. At Ochsner Health, the multimodal model again performed best (c-index: 0.78), surpassing non-image (0.72) and image-only (0.75) models. These findings highlight the benefit of multimodal integration across heterogeneous clinical populations.

### 3.3 Model capability for hazard stratification

Patients were grouped into four risk strata based on predicted hazard scores. Kaplan-Meier survival curves (**Figure 2**) showed that the multimodal model reliably distinguished patients with favorable progression-free survival in the lowest-risk group (1st quartile), with minimal overlap across strata. In contrast, unimodal models displayed curve crossovers and poor separation. Log-rank tests confirmed significant survival differences for the multimodal model (p = 0.04), but not for non-image (p = 0.40) or image-only models (p = 0.21), supporting the superior hazard stratification of the multimodal approach.

### 3.4 Feature Importance analysis

In the multimodal model, several non-image variables contributed strongly to resistance prediction. Based on the magnitude of model coefficients, the most predictive variables included Hispanic ethnicity, Asian ethnicity, and mutations in KIT, KDR, and RB1. To assess statistical significance, we conducted t-tests comparing coefficient values across cross-validation folds against zero. Hispanic ethnicity and RB1 mutation showed significant associations with resistance prediction (p < 0.05), suggesting their robust contribution to model outputs. Full feature statistics are presented in **Table 3**.

To interpret image-based predictions, we generated attention heatmaps from the vision transformer model for two representative patients—one who developed resistance at 7.7 months, and another at 11.2 months. In both cases, high-attention regions (shown in red) corresponded to areas with abnormal nuclear morphology, such as enlarged or irregular nuclei. The model assigned minimal attention to benign or histologically normal regions, indicating its focus on histopathologic features relevant to tumor aggressiveness (**Figure 3**).

Further spatial analysis was conducted using Hover-Net to quantify the cellular composition of histology slides in high- and low-risk groups. The model classified six cell types: tumor, normal, inflammatory, connective tissue, dead, and unclassifiable. As shown in **Figure 4**, high-risk patients exhibited significantly fewer inflammatory cells (p = 0.005), suggesting reduced immune infiltration and possibly a more immunosuppressive microenvironment. Conversely, low-risk patients had a lower proportion of normal cells (p = 0.02), potentially reflecting greater representation of stromal or immune elements. These findings provide biologically meaningful insight into how histologic and cellular features contribute to resistance prediction in the multimodal model.

### 3.5 Performance comparison between model configurations

We conducted an experiment to evaluate how fusion timing affected model performance. All multimodal configurations outperformed unimodal baselines, confirming the benefit of combining data sources. Among the four tested strategies, late fusion—integrating final-layer features from both modalities—achieved the highest mean c-index of 0.81 (**Table 4**).

We also compared different non-image model architectures, including Cox regression, SNN, and FFNN (**Table 5**). The Cox-based model showed the strongest performance (mean c-index: 0.79), while FFNN and SNN performed slightly lower at 0.76 and 0.77, respectively. Given the small sample size, the regularized Cox model provided a favorable balance between performance and model stability.

## 4. Discussion

This study presents a multimodal machine learning model for predicting resistance to EGFR-targeted therapy in NSCLC patients, integrating routinely available histology, genomic, and clinical data. The model achieved strong predictive performance and effective hazard stratification, outperforming unimodal approaches and requiring no manual feature engineering. These findings suggest that multimodal AI may provide a clinically valuable tool for supporting personalized treatment planning and prognosis estimation in patients undergoing EGFR-TKI therapy.

Recent advancements in machine learning, WSI digitization, DNA sequencing, and the adoption of electronic health records have enabled the integration of multimodal AI into clinical workflows for lung cancer management. Currently, treatment decisions rely heavily on expert clinical judgment, which may not be consistently available in low-resource settings. Even in high-resource environments, the interpretation of complex health data often involves considerable uncertainty. Our study addresses this gap by introducing a machine learning tool to predict resistance in EGFR-mutant NSCLC patients receiving osimertinib—an area currently lacking standardized clinical tools. By leveraging data already collected in routine care, our model provides reliable, interpretable risk predictions that could assist clinicians in tailoring treatment strategies and optimizing sequencing of therapies. Moreover, by offering personalized prognosis estimates, the tool may reduce uncertainty for patients, improve quality of life, and support shared decision-making. Given osimertinib's high cost and toxicity profile, this model may help patients weigh expected benefit against financial and health risks more effectively (25).

A key strength of our model is its ability to fuse disparate data types—histopathology, next-generation sequencing, and structured clinical variables—within a single framework. The model achieved a mean c-index of 0.82, consistently outperforming image-only and non-image models across cross-validation folds and across two clinical institutions. Kaplan-Meier curves and log-rank tests confirmed that the multimodal model significantly stratified patients by progression-free survival (p = 0.04), while unimodal models failed to do so. These improvements are likely due to the complementary nature of morphological and molecular features, each capturing different aspects of tumor biology. Furthermore, the model generalized well across both institutions, reinforcing its potential utility in varied clinical settings.

Compared to prior work using radiomics or histology alone for resistance prediction (6,7), our approach demonstrates improved performance and interpretability through multimodal integration. For example, recent studies have applied deep learning to CT imaging or histology to infer mutation status or response but have not combined modalities or evaluated resistance in EGFR-mutant populations (8–10). Our use of late-stage NSCLC patients and inclusion of histology, genomics, and clinical variables represents a novel and practical extension of previous research. Importantly, the model relies only on routinely collected data, offering a pathway for real-world deployment without additional testing or patient burden.

Interpretability analysis revealed key predictors including RB1 mutation and Hispanic ethnicity. RB1 is a tumor suppressor gene implicated in lineage plasticity and histologic transformation, both of which have been associated with resistance to EGFR-TKIs (5,26). Our finding that RB1 mutation significantly contributes to predicted resistance risk aligns with previous work and supports its utility as a molecular risk factor. The association with Hispanic ethnicity, while statistically significant, should be interpreted cautiously due to the small sample size and demographic imbalances. Prior studies have shown mixed findings regarding racial and ethnic disparities in NSCLC outcomes, with some reporting poorer survival among Hispanic and Black patients (27), and others reporting improved outcomes in Hispanic subgroups (28). These inconsistencies emphasize the need for larger, diverse datasets to clarify sociodemographic effects.

Attention-based visualization confirmed that the image model focused on histologic regions rich in deformed nuclei and tumor density—areas typically associated with aggressive disease. Hover-Net-based analysis of nuclear composition further revealed that high-risk patients had significantly fewer inflammatory cells (p = 0.005), while low-risk patients had a higher proportion of stromal and immune elements. These patterns may reflect differences in tumor immune microenvironments, with reduced inflammatory infiltrate potentially indicative of immune evasion or suppressed host response (29).

The primary limitation of this study is the modest cohort size, constrained by stringent inclusion criteria and the availability of multimodal data. This limited sample size informed the use of regularized Cox regression for the non-image branch and motivated a modular, three-stage training pipeline to prevent overfitting. Despite these constraints, the model performed robustly across validation folds and sites. Future work should aim to expand cohort size through multi-site collaborations, enabling validation across more diverse populations and exploration of more complex model architectures, including end-to-end training.

Overall, this study demonstrates the feasibility and value of a multimodal machine learning approach for predicting EGFR-TKI resistance in NSCLC using routinely available clinical data. With further validation, this framework could serve as a clinical decision support tool, informing treatment selection and improving patient counseling—particularly in settings where access to specialized oncology expertise is limited. Broader application of multimodal AI could accelerate precision oncology, enabling more accurate and individualized care across diverse health systems.

## 5. Conclusions

This study presents a multimodal machine learning framework for predicting resistance to EGFR-targeted therapy in NSCLC patients using routinely collected clinical data. By integrating histology images, genomic alterations, and clinical variables, the model achieved strong predictive performance (mean c-index: 0.82) and superior hazard stratification compared to unimodal approaches. It required no manual feature engineering and generalized well across two institutions. The model offers a practical, interpretable tool to support personalized treatment planning, particularly in settings with limited oncologic resources. Key predictors, including RB1 mutation and reduced inflammatory cell presence, were identified as indicators of resistance. These findings underscore the value of combining complementary data modalities to improve prognostic accuracy. Despite the modest sample size, the model remained robust through modular training, regularization, and cross-validation. With further validation in larger, more diverse cohorts, this approach could support real-world clinical decision-making and enhance precision oncology.

## Acknowledgments

We thank John Higgins, MS, for his assistance with data collection for this study and Danielle Cohen, MD, PhD, for her valuable feedback on the manuscript.

## Footnote

**Funding:** This research was supported in part by grants from the US National Library of Medicine (R01LM012837 and R01LM013833) and the US National Cancer Institute (R01CA249758).

**Conflicts of Interest:** The authors declare no competing interests.

**Data Sharing Statement:** The data used in this study are not publicly available due to patient privacy concerns and institutional regulations. De-identified data may be made available from the corresponding author upon reasonable request and with appropriate institutional approvals.

**Ethical Statement:** The study was conducted in accordance with the Declaration of Helsinki (as revised in 2013). Institutional review board (IRB) approval was obtained from both participating institutions: Dartmouth-Hitchcock Medical Center and Ochsner Health. Informed consent was waived due to the retrospective nature of the study and the use of de-identified clinical data. Patient confidentiality was maintained throughout, and all data were handled in accordance with applicable privacy regulations. The authors are accountable for all aspects of the work in ensuring that questions related to the accuracy or integrity of any part of the work are appropriately investigated and resolved.

## Tables

**Table 1.** Baseline demographic, clinical, and treatment characteristics of the cohort, summarized overall and by site. Values are mean (SD) for continuous variables and n (%) for categorical variables.

| Variables | Entire Cohort (n=42) | DHMC (n=23) | Ochsner (n=19) |
|---|---|---|---|
| Age | 69 (12) | 72 (12) | 65 (11) |
| Sex | | | |
| Male | 12 (29) | 6 (26) | 6 (32) |
| Female | 30 (71) | 17 (74) | 13 (68) |
| Race | | | |
| White | 34 (81) | 22 (96) | 12 (63) |
| Black | 4 (10) | 0 (0) | 4 (21) |
| Asian | 4 (10) | 1 (4) | 3 (16) |
| Ethnicity | | | |
| Hispanic | 1 (2) | 0 (0) | 1 (5) |
| Non-Hispanic | 41 (98) | 23 (100) | 18 (95) |
| Prior Immunotherapy | | | |
| No | 39 (93) | 20 (87) | 19 (100) |
| Yes | 3 (7) | 3 (13) | 0 (0) |
| Prior TKI Therapy | | | |
| No | 36 (86) | 17 (74) | 19 (100) |
| Yes | 6 (14) | 6 (26) | 0 (0) |
| Prior Chemotherapy | | | |
| No | 30 (71) | 14 (61) | 16 (84) |
| Yes | 12 (29) | 9 (39) | 3 (16) |
| Prior Surgery | | | |
| No | 34 (81) | 16 (70) | 18 (95) |
| Yes | 8 (19) | 7 (30) | 1 (5) |
| Brain Metastasis | | | |
| No | 24 (57) | 13 (57) | 11 (58) |
| Yes | 18 (43) | 10 (43) | 8 (42) |
| Recurrence | | | |
| No | 28 (67) | 13 (57) | 15 (79) |
| Yes | 14 (33) | 10 (43) | 4 (21) |
| Smoking Status | | | |
| Never Smoker | 23 (55) | 10 (43) | 13 (68) |
| Prior Smoker | 16 (38) | 10 (43) | 6 (32) |
| Current Smoker | 3 (7) | 3 (13) | 0 (0) |
| Smoking Quantity (Packyear) | 13 (22) | 15 (25) | 11 (18) |
| Follow-up Time | 20 (20) | 27 (24) | 12 (8) |
| Number of Histology Slides | 44 | 25 | 19 |
| Number of Pathology Reports | 42 | 23 | 19 |
| Number of Mutations per Patient[c] | 2 (1) | 2 (1) | 2 (1) |

**Table 2.** Comparison of average c-index performance between multimodal and unimodal models across the full cohort and by site. Values represent mean (standard deviation) from five-fold nested cross-validation.

| Modalities | Full Cohort | DHMC | Ochsner |
|---|---|---|---|
| Nonimage modality: Cox regression | 0.77 (0.22) | 0.73 (0.31) | 0.72 (0.30) |
| Image modality: Vision transformer | 0.75 (0.24) | 0.75 (0.35) | 0.75 (0.43) |
| Multimodal (Late fusion) | 0.82 (0.17) | 0.78 (0.30) | 0.78 (0.20) |

**Table 3.** Feature importance based on average Cox regression coefficients from the non-image modality. Values reflect mean coefficients, 95% confidence intervals, and p-values across five cross-validation folds. Statistically significant features (p < 0.05) are highlighted by *.

| Features | Average Coefficient | 95% CI | P-value |
|---|---|---|---|
| Hispanic | 2.77 | 0.84, 4.70 | 0.02* |
| Asian | 2.49 | -3.38, 8.36 | 0.30 |
| KIT | -1.38 | -4.38, 1.60 | 0.27 |
| KDR | 1.34 | -0.38, 3.06 | 0.10 |
| RB1 | 1.14 | 0.03, 2.25 | 0.047* |

**Table 4.** Comparing model performance using different combinations of early and late fusion across image and non-image modalities. Values represent average c-index (standard deviation) from five-fold nested cross-validation.

| Modalities | Unimodal: Nonimage modality | Unimodal: Image modality | Multimodal (late nonimage +late image) | Multimodal (late nonimage +early image) | Multimodal (early nonimage +late image) | Multimodal (early nonimage +early image) |
|---|---|---|---|---|---|---|
| c-index | 0.77 (0.22) | 0.75 (0.24) | 0.82 (0.17) | 0.81 (0.19) | 0.79 (0.18) | 0.81 (0.19) |

**Table 5.** Performance comparison of multimodal models using different non-image architectures (Cox regression, FFNN, and SNN) and fusion strategies. Values represent average c-index (standard deviation) across five-fold nested cross-validation.

| Experiment Settings | | | Results | | | | | |
|---|---|---|---|---|---|---|---|---|
| Non-image modality | Image modality | Fusion layers | Non-image | Image | Multimodal (late nonimage +late image) | Multimodal (late nonimage +early image) | Multimodal (early nonimage +late image) | Multimodal (early nonimage +early image) |
| Cox regression | Transformer | Cox | 0.77 (0.22) | 0.75 (0.24) | 0.82 (0.17) | 0.81 (0.19) | 0.79 (0.18) | 0.81 (0.19) |
| FFNN | Transformer | Cox | 0.70 (0.26) | 0.75 (0.24) | 0.65 (0.23) | 0.79 (0.21) | 0.78 (0.21) | 0.80 (0.19) |
| SNN | Transformer | Cox | 0.76 (0.28) | 0.75 (0.24) | 0.76 (0.23) | 0.79 (0.20) | 0.78 (0.21) | 0.76 (0.22) |

**Figures**

**Figure 1.** Overview of the multimodal model architecture. Histology image features and non-image clinical data, including demographics, genomic alterations, radiology, and treatment history, are processed through separate neural networks. The extracted feature representations are integrated via a fusion layer to generate resistance risk predictions.

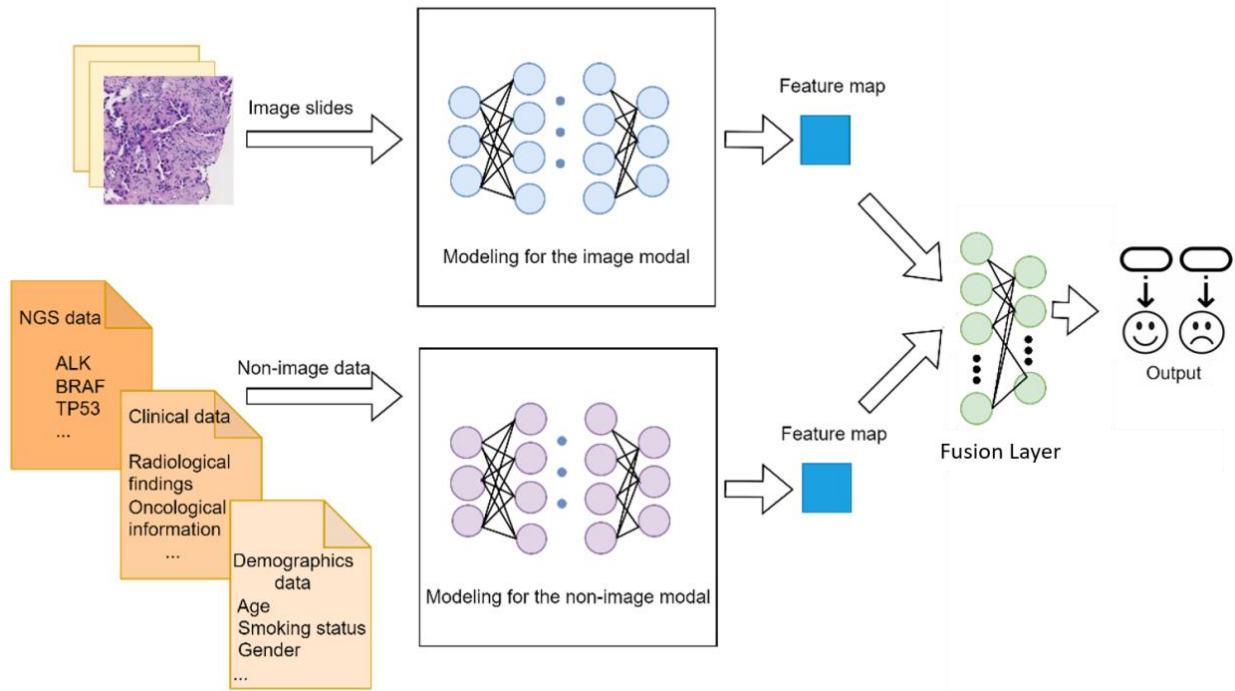

**Figure 2.** Kaplan-Meier curves showing hazard stratification for multimodal, image-only, and non-image-only models. The multimodal model achieved significant separation across risk groups (p = 0.04), outperforming unimodal models.

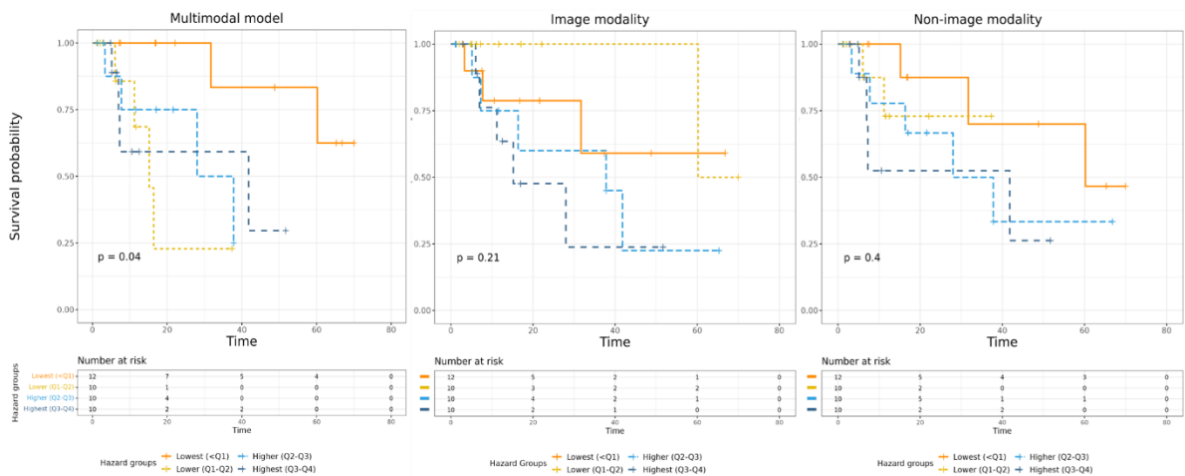

**Figure 3.** Attention map visualization for two representative patients. The model highlights high-risk (H) and low-risk (L) regions within histology slides, focusing on tumor areas with nuclear atypia. Patient A developed resistance at 7.7 months and Patient B at 11.2 months.

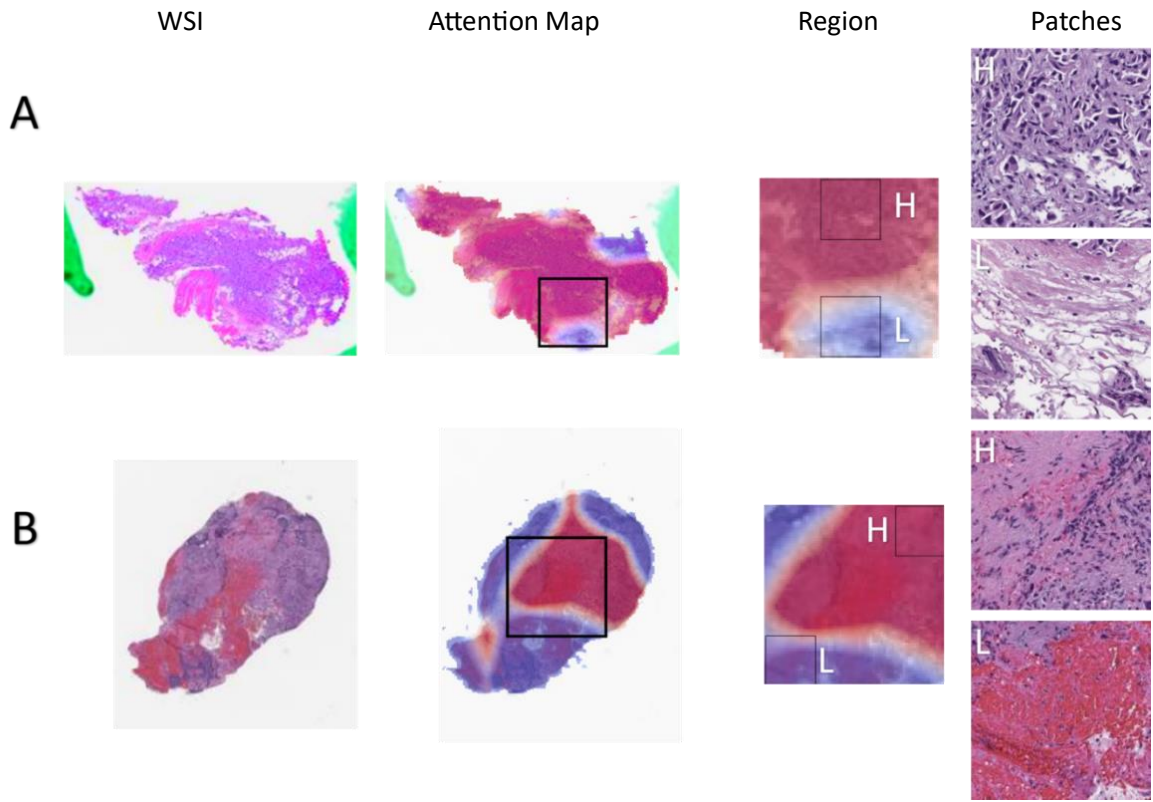

**Figure 4.** Comparison of cell type distributions between predicted high- and low-hazard groups. Patients in the high-hazard group had significantly fewer inflammatory cells (p = 0.005) and more normal cells (p = 0.02) compared to the low-hazard group, based on Mann-Whitney U tests (* indicates statistical significance at p < 0.05).

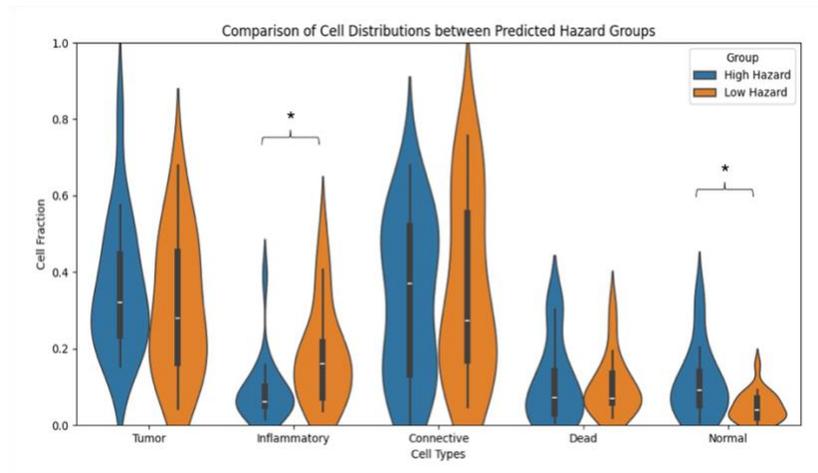